\documentclass[11pt,a4paper]{article}
\usepackage[hyperref]{acl2018}
\usepackage{times}
\usepackage{latexsym}
\usepackage{amsmath}

\usepackage{url}

\aclfinalcopy % Uncomment this line for the final submission
\title{SimpleQuestions Nearly Solved: \\ A New Upperbound and Baseline Approach}

\author{Michael Petrochuk \\
  University of Washington Department \\
  of Computer Science \& Engineering \\
  {\tt mikep5@cs.washington.edu} \\\And
  Luke Zettlemoyer \\
  University of Washington Department \\
  of Computer Science \& Engineering \\
  {\tt lsz@cs.washington.edu} \\}

\date{}

\usepackage{array}
\newcolumntype{L}[1]{>{\raggedright\arraybackslash}p{#1}}

\usepackage{enumerate}
\usepackage[shortlabels]{enumitem}

\newcounter{choice}
\renewcommand\thechoice{\Alph{choice}}
\newcommand\choicelabel{\thechoice.}

  {\list{\choicelabel}%
     {\usecounter{choice}\def\makelabel##1{\hss\llap{##1}}%
       \settowidth{\leftmargin}{W.\hskip\labelsep\hskip 2.5em}%
       \def\choice{%
         \item
       } % choice
       \labelwidth\leftmargin\advance\labelwidth-\labelsep
       \topsep=0pt
       \partopsep=0pt
     }%
  }%
  {\endlist}

  {%
    \setcounter{choice}{0}%
    \def\choice{%
      \refstepcounter{choice}%
      \ifnum\value{choice}>1\relax
        \penalty -50\hskip 1em plus 1em\relax
      \fi
      \choicelabel
      \nobreak\enskip
    }% choice
    \ifvmode\else\enskip\fi
    \ignorespaces
  }%
  {}
  
\begin{document}
\maketitle
\begin{abstract}
    % OBJECTIVE: Strongest hook in the abstract is to explain the solved SimpleQuestions
    % 1. Class of questions is far from solved
    % 2. dataset exists -> study class of questions, it's important as well
    % 3. Nearly solved, with simple methods, why?
    % 3a. We discover a bound
    % 3b. We propose a new model close to the bound
    % 3c. Empirical analysis of remaining errors suggest little room for improvement
    The SimpleQuestions dataset is one of the most commonly used benchmarks for studying single-relation factoid questions. In this paper, we present new evidence that this benchmark can be nearly solved by standard methods. First we show that ambiguity in the data bounds performance on this benchmark at 83.4\%; there are often multiple answers that cannot be disambiguated from the linguistic signal alone. Second we introduce a baseline that sets a new state-of-the-art performance level at 78.1\% accuracy, despite using standard methods. Finally, we report an empirical analysis showing that the upperbound is loose; roughly a third of the remaining errors are also not resolvable from the linguistic signal. Together, these results suggest that the SimpleQuestions dataset is nearly solved.
    % POSSIBLE: Finally, we show the upperbound is loose
\end{abstract}

\section{Introduction}

    % OBJECTIVE: What are we doing?
    We present new evidence that the SimpleQuestions benchmark can be nearly solved by standard methods. First, we show that ambiguity in the data bounds performance; there are often multiple answers that cannot be disambiguated from the linguistic signal alone. Second, we introduce a baseline that sets a new state-of-the-art performance level, despite using standard methods.

    % TODO: Compute the expected accuracy of the upperbound rather than an estimate.
    
    % OBJECTIVE: Explain and expand on "We show that ambiguity in the training data bounds performance on this task at 83.4\%; there are often multiple answers that cannot be disambiguated from the linguistic signal alone."
    Our first main contribution is to show that performance on the SimpleQuestions benchmark is bounded. This benchmark requires predicting a relation (e.g. /film/film/story\_by) and subject (e.g. 090s\_0 [gulliver's travels]) given a question. Consider these examples from the SimpleQuestions dataset:
    \begin{enumerate}[a.]
    \item who wrote gulliver's travels? (film/film/story-\_by, 090s\_0 [gulliver's travels, TV miniseries])
    \item Name a character from gullivers travels (book/-\\book/characters, 0btc7 [gulliver's travels])
    \end{enumerate}
    These examples introduce a fundamental ambiguity. The linguistic signal provides equal evidence for the TV miniseries and the book in both cases, even though only one of the options is labeled as the correct answer.
    We introduce a method for automatically identifying many such ambiguities in the data, thereby producing a new 83.4\% upperbound.
    
    % OBJECTIVE: Explain and expand on "We introduce a baseline that sets a new state-of-the-art performance level at 78.1\% accuracy, despite using standard methods."
    Our second main contribution is a baseline that sets a new state-of-the-art performance level, despite using standard methods. Our approach includes (1) a CRF tagger to determine the subject alias, and (2) a BiLSTM to classify the relation; yielding 78.1\% accuracy for predicting correct subject-relation pairs. 
    
    Finally, we present an empirical error analysis of this model which shows the upperbound is loose and that there is likely not much more than 4\% of performance to be gained with future work on the data. We will publicly release all code and models.

\section{Background}
    % NOTE: Audience is familiar with NLP but not SimpleQuestions or KG directly.
    
    % OBJECTIVE: Introduce the dataset and Freebase with motivation
    Single-relation factoid questions (simple questions) are common in many settings (e.g. Microsoft's search query logs and WikiAnswers questions). The SimpleQuestions dataset is one of the most commonly used benchmarks for studying such questions but remains far from solved. This section reviews this benchmark.
    %Therefore we study SimpleQuestions. Following, for the rest of the section, we provide the reader with background knowledge.
    
    % OBJECTIVE: What is a triple?
    The Freebase knowledge graph (KG) provides the facts for answering the questions in the SimpleQuestions dataset. It includes 3 billion triples of the form (subject, relation, object) (e.g. [04b5zb\_, location/location/containedby, 0f80hy]). We denote such triples as ($s$, $r$, $o$). 
    
    % OBJECTIVE: What is the SimpleQuestions task? w/ example
    % OBJECTIVE: What is a relation and subject?
    % OBJECTIVE: Introduce notation
    % OBJECTIVE: What is the metric for evaluation?
    The SimpleQuestions task is to rewrite questions into subject-relation pairs of the form (subject, relation), denoted in this paper as ($s$, $r$). Each pair defines a graph query that can be used to answer the corresponding natural language question. The subject is a Freebase object with a identifier called an MID (e.g. 04b5zb\_). Freebase objects also typically include one or more string aliases (e.g. mid 04b5zb\_ is named ``fires creek''), which we will use later when computing our upper bounds. The relation is an object property (e.g. location/location/containedby) defined by the Freebase ontology. For example, the question ``which forest is fires creek in'' corresponds with the subject-relation pair (04b5zb\_ [fires creek], location/location/containedby). Finally, the task is evaluated on subject-relation pair accuracy.
    
    % TODO: Worried about not expressing numbers about the dataset like previous research has failed to expand on.
    
    % OBJECTIVE: What is the SimpleQuestions dataset?
    % OBJECTIVE: What is FB2M and FB5M
    The SimpleQuestions dataset provides a set 108,442 simple questions; each question is accompanied by a ground truth triple ($s$, $r$, $o$). This dataset also provides two subsets of Freebase: FB2M and FB5M.\footnote{The FB2M and FB5M subsets of Freebase KG can complete 7,188,636 and 7,688,234 graph queries respectively; therefore, the FB5M subset is 6.9\% larger than the FB2M subset. More previous research has cited FB2M numbers than FB5M; therefore, we report our numbers on FB2M.}
    % OBJECTIVE: Note that we are to use FB2M
    % OBJECTIVE: Note that there is not significant difference between FB2M and FB5M

\section{Dataset Ambiguity and Upperbound} \label{multiple_Interpretations}
    % OBJECTIVE: What are the findings?
    % OBJECTIVE: Example
    Our first main contribution is to show that performance on the SimpleQuestions benchmark is bounded. Consider the question ``who wrote gulliver's travels?'' in Table \ref{table:unanswerable_example}, the linguistic signal provides equal evidence for six subject-relation pairs in the cross product of Table \ref{table:subjects} and Table \ref{table:predicate_to_relation}, including:
    
    \begin{itemize}
    \item (Gulliver's Travels [Book], book/written\_-work/author)
    \item (Gulliver's Travels [TV miniseries], film/-film/written\_by)
    \item \textbf{(Gulliver's Travels [TV miniseries], \\ film/film/story\_by)}
    \end{itemize}
    
    \begin{table}[t!]
    \centering
    \begin{tabular}{|l|l|l|l|}
    \hline \bf Question & \bf Subject & \bf Relation \\ \hline
    who wrote & 090s\_0 & film/film- \\
    gulliver's travels? & & /story\_by  \\
    \hline
    \end{tabular}
    \caption{Unanswerable example from the SimpleQuestions dataset}
    \label{table:unanswerable_example}
    \end{table}
    
    \begin{table}[t!]
    \centering
    \begin{tabular}{|l|l|l|}
    \hline \bf Subject & \bf Description \\ \hline
    0btc7 & Gulliver's Travels (Book) \\
    \textbf{090s\_0} & \textbf{Gulliver's Travels (TV miniseries)} \\
    06znpjr & Gulliver's Travels (American film) \\
    02py9bj & Gulliver's Travels (French film) \\
    \hline
    \end{tabular}
    \caption{FB2M entities with the alias ``gulliver's travels''}
    \label{table:subjects}
    \end{table}
    
    \begin{table}[t!]
    \centering
    \begin{tabular}{|l|l|}
    \hline \bf Relation & \bf Count \\ \hline
    book/written\_work/author  & 132 \\
    film/film/written\_by  & 67 \\
    \textbf{film/film/story\_by} & \textbf{9} \\
    \ldots & \ldots \\
    \hline
    \end{tabular}
    \caption{SimpleQuestions dataset template / predicate ``who wrote $e$?'' relation count}
    \label{table:predicate_to_relation}
    \end{table}
    
    The subject-relation pairs cannot be disambiguated from the linguistic signal alone; therefore, the question is unanswerable. We say a question is unanswerable if there exists multiple subject-relation pairs that are accurate semantic interpretations of the question, as defined in more detail below.

    The ambiguity perhaps comes from annotation process. Annotators were asked to write a natural language question for a corresponding triple ($s$, $r$, $o$). Given only this triple, it'd be difficult to anticipate possible ambiguities in Freebase.
    
    % Link to CODE
    
    % TODO: Reread to see if I missed any citations during the first mention of previous research
    
    \subsection{Approach}
        % OBJECTIVE: General algorithm and intuition for attaining a set of unanswerable questions
        Given an example question $q$ with the ground truth ($s$, $r$, $o$), our goal is to determine the set of all subject-relation pairs that accurately interpret $q$. 
        
        We first determine a string alias $a$ for the subject by matching a phrase in $q$ with a Freebase alias for $s$, in our example yielding ``gulliver’s travels''. We then find all other Freebase entities that share this alias and add them to a set $S$, in our example $S$ is the subject column of Table \ref{table:subjects}.
        
        We define an abstract predicate $p$ (e.g. ``who wrote $e$?'') as $q$ with alias $a$ abstracted. We determine the set of potential relations $R$ (e.g. See Table \ref{table:predicate_to_relation}) as the relations $p$ co-occurs with in the SimpleQuestions dataset.
        
        Finally, if there exists a subject-relation pair $(s, r) \in KG$ such that $r \in R \land s \in S$ we define that as an accurate semantic interpretation of $q$. $q$ is unanswerable if there exists multiple valid subject-relation pairs ($s$, $r$).
        
    \subsection{Results}

        We find that 33.9\% of examples (3675 of 10845) in the SimpleQuestions dataset are unanswerable. Taking into account the frequency of relations for each subject in the KG, we can further improve accuracy by guessing according their empirical distribution, yielding an upperbound of 85.2\%.
        
        Finally, we also found that 1.8\% of example questions ($1587$ of $86755$) in the SimpleQuestions dataset set did not reference the subject. For example ``Which book is written about?'' does not reference the corresponding subject 01n7q (california). We consider these examples as unanswerable, yielding an upperbound of 83.4\%.

\section{Baseline Model}
    Our second main contribution is a baseline that sets a new state-of-the-art performance level, despite using standard methods. Our approach includes (1) a CRF tagger to determine the subject alias, and (2) a BiLSTM to classify the relation.

    \subsection{Approach}
        % OBJECTIVE: Expand on the simple model presented in the introduction... What is the general approach?
        % Give some intuition on the model proposed.
        % What is the model being proposed?
    
        Given a question $q$ (e.g. ``who wrote gulliver's travels?'') our model must predict the corresponding subject-relation pair ($s$, $r$). We predict with a pipeline that first does top-k subject recognition and then relation classification.
        
        We make use of two learned distributions. The subject recognition model $P(a | q)$ ranges over text spans $A$ within the question $q$, in our example including the correct answer ``gulliver’s travels''. This distribution is modeled with a CRF, as defined in more detail below. The relation classification model $P(r|q,a)$ will be used to select a Freebase relation $r$ that matches $q$. The distribution ranges over all relations in Freebase that take objects that have an alias that matches $a$. It is modeled with an LSTM, that encodes $q$, again as defined in more detail below. 
        
        Given these distributions, we predict the final subject-relation pair ($s$, $r$) as follows. We first find the most likely subject prediction according to $P(a|q)$ that also matches a subject alias in the KG. We then find all other Freebase entities that share this alias and add them to a set $S$, in our example $S$ is the subject column of Table \ref{table:subjects}. We define $R$ such that $\forall (s, r) \in KG \{r \in R \land s \in S\}$. Using a relation classification model $p(r | q, a)$, we predict the most likely relation $r_{max} \in R$. 
        
        Now, the answer candidates are subject-relation pairs such that $(s, r_{max}) \in KG \{r \in R \land s \in S\}$. In our example question, if $r_{max}$ is film/film/story\_by then $S$ is both subjects 06znpjr (Gulliver’s Travels, American film) and 02py9bj ( Gulliver’s Travels, French film). Because there is no explicit linguistic signal to disambiguate this choice, we pick the subject $s_{max}$ that has the most facts of type $r_{max}$.
        
    \subsection{Model Details}
        % OBJECTIVE: How did we set up our model in order to discover the end-to-end results?
        % End-To-End:
        % - What data did we use?
        % - What did we evaluate on?
        % - How were our models trained?
        Our approach requires two models, in this section we cover training and configuring these models. Note we train and configure the model on the SimpleQuestions 75,910 training examples and 10,845 validation examples respectively.
            
        \paragraph{Top-K Subject Recognition}
            % OBJECTIVE: Summarize this section
            % Use approach notation
            We model top-k subject recognition $P(a | q)$ using a linear-chain conditional random field tagger (CRF) with a conditional log likelihood loss objective. k candidates are inferred with the top-k viterbi algorithm.
        
            % OBJECTIVE: dataset
            Our model is trained on a dataset of question tokens and their corresponding subject alias spans using IO tagging. The subject alias spans are determined by matching a phrase in the question with a Freebase alias for the subject.
            
            % OBJECTIVE: Talk about a couple of the hyperparameters
            As for hyperparameters, our model word embeddings are initialized with GloVe and frozen. Adam, initialized with an learning rate of 0.0001, is employed to optimize the model weights. Finally, we halve the learning rate if the validation accuracy has not improved in 3 epochs.
            
            % OBJECTIVE: hyperparameter tuning
            All hyperparameters are hand tuned and then a limited set are further tuned with grid search to increase validation accuracy. In total we evaluated at most 100 hyperparameter configurations.
            
        % OBJECTIVE: Talk about the models used to execute on the approach...
        \paragraph{Relation Classification}
            % OBJECTIVE: Summarize this section
            % Use approach notation
            Relation classification $P(r | q,a)$ is modeled with a one layer BiLSTM batchnorm softmax classifier that encodes the predicate $p$ and uses a negative log likelihood loss objective. We define an abstract predicate $p$ (e.g. ``who wrote $e$?'') as $q$ with alias $a$ abstracted.
            
            The model is trained on a dataset of predicate $p$ and relation set $R$ to ground truth relation $r$. These values are attained by following our approach in Section 4.1 until the values are declared.
            
            As for hyperparameters, the model word embeddings are initialized with FastText~\cite{bojanowski2016enriching} and frozen. The AMSGrad variant of Adam~\cite{j.2018on}, initialized with an learning rate of 0.0001, is employed to optimize the model weights. Finally, we double the batch size~\cite{Smith2017DontDT} if the validation accuracy has not improved in 3 epochs.
            
            All hyperparameters are hand tuned and then a limited set are further tuned with Hyperband~\cite{Li2017HyperbandBC} to increase validation accuracy. Hyperband is allowed at most 30 epochs per model and a total of 1000 epochs. In total we evaluated at most 500 hyperparameter configurations.

    \subsection{Results}
        % OBJECTIVE: What were the results of experimenting with our model?
        
        Following running our model on the SimpleQuestions 21,687 test set examples, we present our results on the SimpleQuestions task. Note we run on the test set only once to measure generalization. 
    
        \paragraph{SimpleQuestions Task}
        % - OBJECTIVE: How did the end-to-end results work out? *Compared to other papers
        
        \begin{table}[t!]
        \centering
        \begin{tabular}{|l|l|}
        \hline \bf Previous Work & \bf Acc. \\ \hline
        Random guess~\cite{Bordes2015LargescaleSQ} & 4.9 \\
        \hline
        Memory NN~\cite{Bordes2015LargescaleSQ} & 61.6 \\
        \hline
        Attn. LSTM~\cite{He2016CharacterLevelQA} & 70.9 \\
        \hline
        GRU ~\cite{Lukovnikov2017NeuralNQ} & 71.2 \\
        \hline
        BiGRU-CRF \& BiGRU &  73.7 \\
        ~\cite{Mohammed2017StrongBF} &  \\
        \hline
        BiLSTM \& BiGRU & 74.9  \\
        ~\cite{Mohammed2017StrongBF} & \\
        \hline
        BiGRU \& BiGRU~\cite{Dai2016CFOCF} & 75.7 \\
        \hline
        CNN \& Attn. CNN \&  &  76.4 \\
        BiLSTM-CRF~\cite{DBLP:journals/corr/YinYXZS16} & \\
        \hline
        HR-BiLSTM \& CNN \& & 77.0 \\
        BiLSTM-CRF ~\cite{Yu2017ImprovedNR} &  \\
        \hline
        \bf BiLSTM-CRF \& BiLSTM (Ours) & \bf 78.1 \\
        \hline
        \end{tabular}
        \addtocounter{footnote}{-1}
        \caption{Summary of past results on the SimpleQuestions benchmark along with the neural models employed. Note that an ``\&'' indicates multiple neural models.}
        \label{table:results}
        \end{table}
        
        Our baseline model achieves 78.1\% accuracy, a new state-of-the-art without ensembling or data augmentation (Table \ref{table:results}). 
        %Note we only list results attained by training on the SimpleQuestions training dataset and without using ensembles.
        % TODO: Ask about "Due to the page limit"
        These results suggest that relatively standard architectures work well when carefully tuned, and approach the level set by our upper bound earlier in the paper. 
        
        %and tuning models toward the problems affects performance significantly. For us, this meant tuning hyperparameters as well feature engineering like picking an optimal text normalization strategy to index KG aliases. Due to the page limit, we do not follow up with an ablation study to formally quantify the affect of tuning individual data processing modules.
        
        % \stepcounter{footnote}\footnotetext{Evaluated on FB5M.}
        % \stepcounter{footnote}\footnotetext{See footnote 2.}
        \stepcounter{footnote}\footnotetext{\citealt{Tre2017NoNT} reported a 86.8\% accuracy but we and \citealt{Mohammed2017StrongBF} have not been able to replicate their results. \citealt{Wang2017QuestionAW} scored 77.5\% but removed 0.5\% of the test examples.}
        
        \paragraph{Further Qualitative Analysis}
        
        We also analyze the remaining errors, to point torward directions for future work.
        
        In Section \ref{multiple_Interpretations}, we showed that questions can provide equal evidence for multiple subject-relation pairs. To remove this ambiguity, we count any of these options as correct, and our performance jumps to 91.5\%. 
        
       The remaining 8.5\% error comes from a number of sources. First, we find that 1.9\% of examples were incorrect due to noise mentioned in Section \ref{multiple_Interpretations}. Finally, we are left with a 6.5\% gap. To understand the gap, we do an empirical error analysis on a sample of 50 negative examples.
        
        % TODO: Reword this better
        First we found that for 14 of 50 cases the question provided equal linguistic evidence for both the ground truth and false answer, similar to the dataset ambiguity found in Section \ref{multiple_Interpretations}, suggesting that our upper bound is loose. We note that Section \ref{multiple_Interpretations} did not cover all possible question-subject-relation pair ambiguities. The approach relied on exact string matching to discover ambiguity; therefore, missing similar paraphrases. For example, the abstract predicate ``what classification is $e$'' had more examples than ``what classification is \textbf{the} $e$'' allowing our approach to programmatically define more subject-relation pair ambiguities for the former predicate than the later.
        
        The remaining 36 of 50 cases were linguistic mistakes by our model. Among the 36 cases, we identified these error cases:
        \begin{itemize}
          \item \textbf{Low Shot} (16 of 36) The relation was seen in the training data less than 10 times.
          \item \textbf{Subject Span} (14 of 36) The subject span was incorrect.
          \item \textbf{Noise} (2 of 36) The question did not make grammatical sense.
        \end{itemize}
        
        Finally, the error analysis of this model shows that the upperbound is loose. There is likely not much more than 4\% of performance to be gained with future work on the data.
        
        % TODO: Talk to Luke about adding a related work section to emphasize the key differences in the pipeline

\section{Conclusions and Future Work}

The SimpleQuestions dataset is one of the most commonly used benchmarks for studying single-relation factoid questions. In this paper, we presented new evidence to suggest that this benchmark can be nearly solved by standard methods. These results suggest there is likely not much more than 4\% to be gained with future work on the data. 

\bibliography{acl2018}
\bibliographystyle{acl_natbib}

\end{document}